\documentclass[conference]{IEEEtran}
\IEEEoverridecommandlockouts
\usepackage{svg}   

\usepackage{booktabs}
\usepackage{makecell}

\newcommand{\approach}{\textsc{ODD-Tax-232}}

\usepackage{blindtext}

\usepackage{cite}
\usepackage{amsmath,amssymb,amsfonts}
\usepackage{algorithmic}
\usepackage{graphicx}
\usepackage{textcomp}
\usepackage{xcolor}

\usepackage[colorlinks=true, linkcolor=black, citecolor=black, urlcolor=black]{hyperref}
\usepackage{cleveref}

\crefname{section}{Sec.}{Secs.}
\crefname{figure}{Fig.}{Figs.}
\crefname{table}{Tab.}{Tabs.}

\def\BibTeX{{\rm B\kern-.05em{\sc i\kern-.025em b}\kern-.08em
    T\kern-.1667em\lower.7ex\hbox{E}\kern-.125emX}}
\begin{document}

\title{Operating Within the Operational Design Domain: Zero-Shot Perception with Vision-Language Models}

\author{Berkehan Ünal$^{1,2,3}$, Hauke Dierend$^{1,4}$, Dren Fazlija$^{2}$, and Christopher Plachetka$^{1,5}$%

\thanks{$^{1}$Volkswagen Aktiengesellschaft, Wolfsburg, Germany {\tt\small \{forename.surname\}@volkswagen.de}}
\thanks{$^{2}$L3S Research Center, Leibniz University Hannover, Hanover, Germany. {\tt\small \{forename.surname\}@L3S.de}}
\thanks{$^{3}$Faculty of Information Technology, University of Jyväskylä, Jyväskylä, Finland. berkehan.u.unal@jyu.fi}
\thanks{$^{4}$MOIA GmbH, Hamburg, Germany. hauke.dierend@moia.io}
\thanks{$^{5}$Motor AI GmbH, Berlin, Germany. christopher.plachetka@motor-ai.com}
}
%\author{Anonymous Authors}

\maketitle

\begin{abstract}
Over the last few years, research on autonomous systems has matured to such a degree that the field is increasingly well-positioned to translate research into practical, stakeholder-driven use cases across well-defined domains. 
However, for a wide-scale practical adoption of autonomous systems, adherence to safety regulations is crucial. 
Many regulations are influenced by the Operational Design Domain (ODD), which defines the specific conditions in which an autonomous agent can function. 
This is especially relevant for Automated Driving Systems (ADS), as a dependable perception of ODD elements is essential for safe implementation and auditing. 
Vision–language models (VLMs) integrate visual recognition and language reasoning, functioning without task-specific training data, which makes them suitable for adaptable ODD perception. 
To assess whether VLMs can function as zero-shot "ODD sensors" that adapt to evolving definitions, we contribute $(i)$ an empirical study of zero-shot ODD classification and detection using four VLMs on a custom dataset and Mapillary Vistas, along with failure analyses; 
$(ii)$ an ablation of zero-shot optimization strategies with a cost–performance overview; 
and $(iii)$ a suite of reusable prompting templates with guidance for adaptation. 
Our findings indicate that definition-anchored chain-of-thought prompting with persona decomposition performs best, while other methods may result in reduced recall. 
Overall, our results pave the way for transparent and effective ODD-based perception in safety-critical applications.
\end{abstract}


\section{Introduction}

\noindent With the rapid pace of AI innovations, manufacturers and end-users desire more autonomous robotic systems. 
%Although researchers have successfully employed existing AI paradigms to develop these systems, their applicability primarily remains limited to virtual environments.
Although real-world deployments are growing, translating this demand into widespread adoption depends less on capability than on verifiable safety.
The binding constraint is compliance with stringent regulatory standards set by public authorities.
A pivotal concept in such regulations is the Operational Design Domain (ODD), which SAE J3016~\cite{sae2014odd} defines as the operational conditions under which an autonomous system is intended to function safely. 
%\TODO{ -> A pivotal concept in such regulations is the Operational Design Domain (ODD), which delineates the operational conditions under which an autonomous system is intended to function safely~\cite{mendiboure2023odd}.  (So ist es ofizieller: SAE J3016 explicitly defines ODD as the operating conditions )}
%Applications of ODDs include the autonomous deployment of maritime systems~\cite{nakashima2025maritime}, agricultural robots~\cite{happich2025agriculturalindustryinitiativesautonomy}, aviation systems~\cite{torens2024operational}, trains~\cite{meng2021trains}, and mining systems~\cite{castellanos2022context}.
%, and orbital robotics~\cite{lampariello2025validation}.  
%\blindtext

ODDs have been applied across domains such as maritime and aviation systems~\cite{nakashima2025maritime, torens2024operational}, with autonomous driving at the forefront of regulatory development.

However, at the forefront of ODD-based safety regulations are autonomous driving systems (ADS). With a projected revenue of \$300 billion to \$400 billion by 2035~\cite{mckinsey2023autonomous}, these systems have substantial economic potential, while being widely discussed in the public media landscape ~\cite{Matin2024publicperception}. 
As such, our contributions to ODD-based technologies will focus on this specific use case.
In ADS, the ODD comprises stationary elements (e.g., road type, geography, permanent signage), transient elements (e.g., weather, illumination, traffic conditions), and moving elements (e.g., pedestrians) as structured in established ODD taxonomies \cite{ISO34503_2023, BSI2020PAS1883}.
Ensuring compliance with the ODD is a regulatory requirement in many markets, like Germany~\cite{kba2023}, and a technical necessity to prevent system activation under unsuitable conditions ~\cite{ISO21448_2022}. 
%\TODO{ -> ADS providers must not deploy an ADS outside of its ODD, since operating beyond these defined conditions is unsafe. (Das Statement ist nur eine Halbwahrheit, würde das weglasen); Dren: Was machen wir dann mit den nächsten 2 Sätzen darunter? -> Accordingly, the ADS should engage only within its declared ODD and, when encountering conditions such as unmarked roads where reliable performance is not ensured, either block engagement or promptly initiate a safe fallback (e.g., transition to the driver or a minimal-risk manoeuvre).}
Accordingly, the ADS should engage only within its declared ODD and, when encountering conditions such as unmarked roads where reliable performance is not ensured, either block engagement or promptly initiate a safe fallback (e.g., a minimal-risk maneuver)~\cite{ISO21448_2022,gyllenhammar2020towards}. Furthermore, roads with environmental conditions that exceed the system’s capabilities should be excluded from the operational network altogether~\cite{gyllenhammar2020towards}.
% For example, suppose a system struggles to perform reliably in situations without road marking.
% In that case, it should either refrain from entering such areas or require the driver to operate the vehicle there manually. 

% Figure

%While we can manage some static ODD taxonomy elements through geofencing~\cite{Dimino1999}, more dynamic elements, such as sudden dense fog, necessitate an on-site ODD taxonomy elements perception system.

The effective assessment of ODD compliance requires a perception system capable of detecting and differentiating between a vast array of environmental conditions. Such a system must remain flexible to the addition of new concepts to the ODD taxonomy to support evolving safety requirements.

To address this, we require a robust computer vision system capable of identifying a wide range of concepts, ranging from pavement marking types to environmental road states.

%Since the manual development and training of dedicated perception models for each of these diverse conditions is associated with extremely high effort, we investigate the potential of vision-language models (VLMs). Due to their zero-shot capabilities, VLMs offer a flexible and efficient alternative for capturing complex ODD scenarios without the need for specialized training data, thereby accelerating the validation of automated systems.

%Given their strong understanding of detailed textual descriptions and ability to process images without propriety training, we believe that VLMs can lead to a new era of dynamic ODD taxonomy element perception. 

Since the manual development and training of dedicated perception models for each of these diverse conditions is associated with extremely high effort, we investigate the potential of vision-language models (VLMs). By integrating visual recognition and language reasoning, VLMs can process images without task-specific training data, offering a flexible and efficient alternative for capturing complex ODD scenarios, thereby accelerating the validation of automated systems and opening a new era of dynamic ODD taxonomy element perception.

Specifically, these models have the potential to directly support real-world ADS safety workflows by enabling continuous ODD compliance monitoring from on-board perception, particularly for rare or evolving conditions that are impractical to exhaustively pre-map. Zero-shot ODD concept recognition further allows safety engineers to rapidly prototype, validate, and stress-test ODD boundaries in simulation without retraining perception models, reducing development and verification overhead. Finally, the interpretability and prompt-level configurability of VLM-based systems can facilitate certification-relevant analyses by making ODD assumptions explicit, auditable, and adaptable to regulatory updates.

To investigate this hypothesis and to lay the foundation for future research into VLM-based solutions for ODD-abiding perception, our work offers the following contributions:
$(i)$ an experimental study comparing the zero-shot classification and detection abilities of different VLMs, ranging from large, closed-source, general-purpose models, to smaller open-sourced specialized systems, and analyzing their failure modes and common patterns;
$(ii)$ an in-depth analysis of zero-shot optimization strategies for ODD perception, including a cost-performance trade-off characterization across token budgets (cf.~\cref{fig:tradeoff}), and distilled insights that transfer to other ODD applications;
$(iii)$ a publicly-available suite of prompting templates, accompanied by adaptation guidelines and usage notes to facilitate reuse and extension by other researchers (available at~\url{https://github.com/Berkehanunal/odd-tax-232}).

% Figure
\begin{figure*}[t!]
  \centering
  %\includesvg[width=\linewidth]{figures/Group}%
  \includegraphics[width=0.8\linewidth]{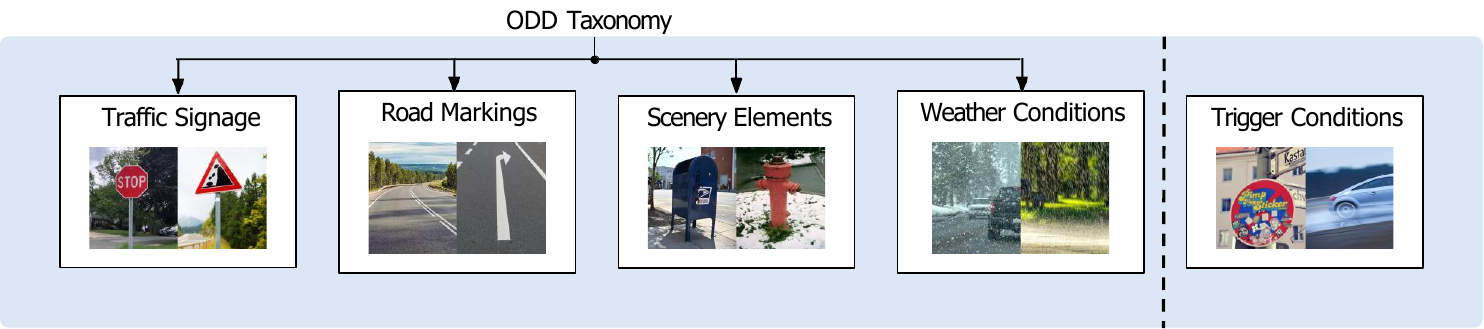}
    \caption{High-level overview of our custom dataset~\approach{}. All 232 concepts either represent Traffic Signage (27/232), Road Markings (55/232), Scenery Elements (96/232), Weather Conditions (12/232), or Trigger Conditions (42/232). Here, \emph{Trigger Conditions} refer to system-independent environmental properties (e.g., decreased sign legibility) rather than the system-dependent initiators defined in ISO~21448~\cite{ISO21448_2022}. Each image is sourced from public platforms and depicts the annotated concept unambiguously.}
  \label{fig:taxonomy}
\end{figure*}

\section{Foundational Concepts \& Related Work}

\subsection{Operational Design Domains for Autonomous Driving}

\noindent ADS are integrated technological platforms that enable vehicles to perform driving tasks with reduced or no human intervention~\cite{hancock2019future} by combining advanced sensors, control algorithms, and AI to perceive the environment, make decisions, and execute maneuvers. 
While extensive public testing has demonstrated that current ADS can navigate complex real-world scenarios~\cite{broggi2015proud}, the safe implementation of ADS requires a robust safety assurance framework that is recognized by authorities, end-users, regulatory bodies, and manufacturers.
The ODD assumes this pivotal role, playing a central part in defining system boundaries.
Engineers employ ODD concepts extensively to derive design requirements, generate test scenarios, and conduct safety and risk analyses throughout the development and deployment process~\cite{gyllenhammar2020towards}.
% \TODO{
% Since SAE J3016 was first published in 2014 (using ``driving mode''), subsequent revisions formalized the term ODD, which has since been widely adopted by industry and regulators~\cite{sae2014odd}. Dieser Satz kann eigentlich weggelassen werden }

This wide-scale adoption has led to the emergence of many so-called ODD taxonomies~\cite{BSI2020PAS1883, ISO34505_2025}.
Such taxonomies represent hierarchical element definitions organized as a formalized tree structure, where high-level nodes describe general concepts, while lower-level branches capture increasingly specific environmental and operational features. 
Common taxonomy categories include weather conditions, road infrastructure, traffic participants, connectivity, and vegetation. 
These structured classifications serve as a shared vocabulary for describing the operational contexts in which an ADS is expected to function.

In this work, we systematically describe driving scenes through a subset of the ODD taxonomy defined in ISO~34503~\cite{ISO34503_2023}, which we extended with additional concepts from ADS development to provide a more granular characterization of complex operational scenarios.

% In this work, we systematically describe driving scenes through the ODD taxonomy defined by Richter et al.~\cite{odd_taxonomy2025}. 
% The taxonomy comprises more than 1,000 unique concepts, representing, to the best of our knowledge, the most fine-grained ODD taxonomy available. It achieves this by integrating existing taxonomies, such as ISO 34503 ~\cite{ISO34505_2025}, and extending them with domain knowledge derived from ADS development.

% \TODO{Problem das dataset was wir benutzen ist nur ein Subset, von dahher müsste man das Orginal ODD dataset anders beleuchten.  In this work, we systematically describe driving scenes through a subset of the ODD taxonomy defined by Richter et al.~\cite{odd_taxonomy2025}, which integrates existing taxonomies such as ISO 34503 ~\cite{ISO34505_2025}, and extends them with domain knowledge derived from ADS development.}

\subsection{Zero-Shot Optimization Strategies}\label{sec:zeroshotstrategies}
\noindent One of the most intriguing aspects of large language models (LLMs) and VLMs is their ability to perform well on new tasks without any fine-tuning of the model weights. 
Instead, one can optimize models for specific tasks using strategies like prompt engineering and retrieval-augmented generation~\cite{10.1145/3560815}.

Prompt engineering involves crafting inputs to elicit accurate and coherent responses from models. 
A basic method is zero-shot prompting, where the model is given only a task instruction, without any demonstrations or labeled examples, and must rely entirely on its pre-trained knowledge to generate an appropriate response~\cite{NEURIPS2020_1457c0d6}.

Enhancements like Chain-of-Thought (CoT) prompting encourage models to outline intermediate reasoning steps, improving performance on complex tasks~\cite{10.5555/3600270.3602070}. 
Variants such as Auto-CoT~\cite{zhang2022automatic} and self-consistency~\cite{wang2023selfconsistencyimproveschainthought} further strengthen CoT prompting.
Beyond CoT prompting, researchers repeatedly demonstrated that invoking so-called personas within a language model can lead to performance boosts in both language- and image-based tasks~\cite{tseng-etal-2024-two,choi2024beyond,luz2025helpful,salewski2023context}. 
These prompts can range from simple role assignments (e.g., ``You are a supportive high school physics teacher") to detailed behavioral personas involving stylistic constraints, tone modulation, or domain-specific expertise. 
We refer to Tsang et al.~\cite{tseng-etal-2024-two} for a detailed breakdown of persona prompting strategies.

In addition to optimizing prompt design, it is possible to enhance LLMs and VLMs by integrating them with knowledge bases containing domain-specific information. 
This approach is known as retrieval-augmented generation (RAG), which combines the advantages of both parametric and non-parametric memory systems~\cite{lewis2020retrieval}. 
% In this method, a neural retriever dynamically fetches relevant documents from a dense index, such as Wikipedia. 
% A pre-trained sequence-to-sequence generator then utilizes these documents to produce the final output. 
Researchers have demonstrated that the performance of RAG pipelines can be improved by utilizing advanced prompting strategies, such as Chain-of-Verification (CoV)~\cite{dhuliawala2024chain} and Chain-of-Note (CoN)~\cite{yu2024chain}.

In this work, we investigate the performance gains for ODD condition detection by VLMs obtained through the strategies discussed.

\subsection{Vision-Language Models for Autonomous Driving}

\noindent VLMs extend conventional perception in autonomous driving by combining visual evidence with language-guided reasoning. Prior work has successfully leveraged VLMs for high-level environmental context detection. ContextVLM~\cite{10920066}, for example, introduced a framework for identifying 24 driving-relevant contexts, such as weather and lighting conditions. Albeit effective, its scope is limited to coarse scene attributes. While high-level context detection has seen steady progress, practical autonomy increasingly demands fine-grained recognition of ODD taxonomy concepts, including road geometry, markings, surface conditions, and traffic control devices. Our work advances this direction by targeting a broader, more granular ODD taxonomy with explicit spatial grounding. This objective aligns with benchmarks such as VLADBench~\cite{li2025vladbench}, which emphasize ``Geometric and Element Recognition,'' and position fine-grained categorization as a distinct capability beyond scene-level tagging. Notably, recent findings in adjacent domains (e.g., augmented reality) report that VLMs struggle with precise state recognition and attribute granularity, reflected in low F1 scores for fine-grained tasks~\cite{smith2024finegrainedvl}. These observations underscore the challenges of our setting and motivate the specialized pipeline and zero-shot optimization strategies we pursue.

A growing body of surveys systematizes how foundation models, LLMs, VLMs, and related architectures, integrate with perception, prediction, and planning~\cite{hou2024vlmADASurvey, gao2025foundation}. These works map tasks, metrics, and datasets, and articulate open challenges across the autonomy stack. In contrast, we concentrate on \emph{fine-grained ODD element extraction} with explicit spatial grounding and georeferencing, operationalizing VLM outputs for ODD mapping at scale. Complementary lines of work leverage generative and multimodal models to synthesize diverse scenarios for robustness and testing (e.g., OmniTester~\cite{lu2024omnitester}, WEDGE~\cite{marathe2023wedge}). Our contribution is orthogonal: we produce structured, georeferenced ODD data that can enrich upstream perception in adverse conditions, seed simulation assets, and support proactive risk analysis. Relatedly, SafeAuto~\cite{zhang2025safeautoknowledgeenhancedsafeautonomous} applies retrieval-augmented generation to encode safety knowledge from prior experience; our pipeline provides foundational, structured ODD signals that such knowledge systems could readily consume.

\section{Experiment Design}

\subsection{Optimizing Zero-Shot Perception}\label{sec:prompts}
\noindent To establish a fair and accurate baseline for later comparisons of different VLMs, we first need to conduct zero-shot optimization. 
For this purpose, we explore various strategies described in~\cref{sec:zeroshotstrategies}, using a fixed model and a newly created custom dataset.
%In this experiment, we created a custom dataset that includes 10 samples for each taxonomy element, resulting in a total of 2,320 images.
We then analyze the effects of different optimization strategies on the GPT-4o model~\cite{hurst2024gpt}.
%\TODO{Wann gibt man seinen Datasets coole Namen?}

\textbf{Custom ODD-Taxonomy Dataset.} We curated a custom ODD-Taxonomy dataset (\approach{}) comprising 232 concepts (10 samples per concept, totaling 2,320 images). With 10 samples per concept, ODD-TAX-232 represents a deliberate starting point: each image was manually curated to depict a concept, prioritizing annotation quality over quantity.
These elements (see~\cref{fig:taxonomy} for examples) were selected due to their known impact on autonomous vehicle (AV) subsystems, such as perception or scene understanding, which can potentially affect safety. Moving elements, such as pedestrians and vehicles, were excluded as their detection is well-served by existing specialized detectors, and their dynamic nature makes static image classification an unsuitable evaluation paradigm.
%they represent complex ODD taxonomy concepts, are underrepresented in non-autonomous driving datasets, and include conditions (e.g., weather and triggering factors \TODO{hier trigger conditions rausnehmen und stattdessen weather, signage occlusion/quality issues, complex junctions, and surface contamination such as snow or debris}) that frequently occur in AV scenarios.
% To avoid ambiguity in the classification task, we manually ensured that each image represents exactly one taxonomy concept.
While each image may contain more than one taxonomy concept, we manually attribute each image to the predominant concept to minimize ambiguity in the classification task. Such an approach restricts our evaluation to recall (based on true positive) detections only, but drastically reduces manual annotation efforts. The images were sourced from public domains, adhering to strict criteria for visibility and diversity. To ensure geographical relevance, the dataset focuses on infrastructure and signage from the European Union, the United Kingdom, and the USA. The imagery primarily captures daytime conditions (e.g., midday) across a balanced mixture of urban and rural environments to reflect common operational scenarios. The dataset and prompting templates are publicly available at~\url{https://github.com/Berkehanunal/odd-tax-232}.

%Regarding potential dataset biases, we acknowledge the reliance on public platforms such as Wikimedia Commons. While these sources provide high-quality, unambiguous depictions of specific ODD concepts, they may over-represent clear visibility conditions compared to raw on-vehicle sensor data. However, we mitigate this by specifically including categories for "Trigger Conditions" that account for quality issues and occlusions. To ensure high annotation quality, the dataset underwent a multi-stage validation process: an initial selection by a single annotator was subsequently reviewed and corrected by two independent ODD experts, ensuring high inter-annotator reliability and alignment with regulatory taxonomy definitions.

Our dataset is primarily designed for classification annotations, facilitating structured prompting and precise measurements. Unlike larger benchmarks,~\approach{} aims at fine-grained environmental features, supporting evaluations of VLM recognition performance with regard to ODD taxonomy concepts, with prospects of further advancements to bounding box detection tasks.

\textbf{Prompt Design Rationale.}
In all subsequent experiments, we define the model as a ``professional scene understanding system for AV safety analysis" and use motivational framing to encourage cautious, high-certainty predictions. 
Similar to Salewski et al.~\cite{salewski2023context}, we use a reward-penalty structure that discourages overconfident guessing by providing rewards for correct detections and penalties for incorrect ones.
Furthermore, we explicitly define behavioral constraints: responses must be confident, accurate, and structured.
We also provide a fixed terminology for ODD taxonomy concepts to prevent uncontrolled variation. 
We task the model with outputting JSON-structured data, which we evaluate using recall against the ground-truth labels. 
% In all subsequent experiments, we establish a clear role for the model by defining it as a "professional scene understanding system trained to assist autonomous vehicle safety analysis". 
% Furthermore, we utilize motivational framing to nudge the model toward cautious, high-certainty predictions. 
% The inclusion of a simulated reward-penalty structure ("You will receive a financial reward for every correct detection [...] You will incur a severe penalty for any incorrect, imagined, or hallucinated detection." ) discourages overconfident guessing~\cite{salewski2023context}.
% Third, the prompt explicitly enumerates behavioral constraints that the model must follow: responses must be confident, accurate, and structured.
% Finally, the prompt includes a predefined and exhaustive taxonomy of ODD elements.
% This closed-world design forces the model to select from a fixed list of valid classes, thereby preventing uncontrolled variation and lexical drift.
% %In doing so, the prompt ensures alignment with task-specific ontologies used in the development and evaluation of autonomous vehicles.
% Additionally, models are tasked to generate JSON–structured output, which we evaluate using recall against the ground-truth labels.
\vfill

\textbf{Instructional Zero-Shot: Baseline \& Label Normalization.}
Some of our crafted prompts merely change how the task is framed rather than the model's assessment procedure. For one, we have our baseline prompt (referred to as \texttt{Flat Taxonomy}), where we define the entire label space as a single list without any semantic grouping. Beyond the vanilla prompt, we also investigated label aliasing, in which we map ambiguous taxonomy concepts into explicit, action-oriented descriptors while preserving a deterministic mapping back to the canonical taxonomy. For instance, our aliasing maps the generic term "number" to the more informative description "painted\_speed\_or\_route\_number". 
We employ label aliasing on the persona-based prompt described below (\texttt{Persona + Label Aliasing + Taxonomy}). 

\textbf{Process-Oriented Zero-Shot: Self-Refinement, Two-Stage \& Personas.}
% Many of our assessed prompts actively alter the process the model follows, rather than the phrasing of the prompt.
Many of our assessed prompts do more than rephrase instructions; they restructure the model's inference pipeline.
For instance, the \texttt{Reevaluate} prompt asks the model to reassess its previous predictions, verifying if each label is supported by visual evidence in the image.
Another two-stage process is our \texttt{Road-dependent} prompt, where we ask the model to $(i)$ classify the street type in the image and $(ii)$ classify visible concepts conditioned on that type, restricting predictions to context-appropriate labels.
Finally, we invoke latent personas within VLMs through our \texttt{Persona Decomp.} prompt, where we decompose the task into specialized sub-prompts, each handled by a dedicated domain expert. 
Concretely, we partition the taxonomy into five domain experts: a Scenery Expert (e.g., trees, buildings, barriers), a Sign Expert, a Trigger Condition Expert, a Weather Expert, and a Markings Expert (see~\cref{fig:taxonomy} for sample images of each category).
Each sub-prompt supplies its expert with a scoped label list and precise, task-specific instructions.

\textbf{Zero-Shot with Retrieval (RAG).}
To minimize the token budget of our persona-based prompt, we investigate the use of RAG. In the resulting prompt (\texttt{Persona + RAG}), the VLM initially produces a scene description of the given image. 
The description is then used to query a knowledge base comprising textual descriptions of all 232 ODD concepts. 
The corresponding embeddings are provided by the all-MiniLM-L6-v2 model\footnote{\url{https://huggingface.co/sentence-transformers/all-MiniLM-L6-v2}} from the SentenceTransformers library. 
Finally, we task the VLM to detect ODD elements based on the image and the eight most relevant retrieved textual descriptions.

\begin{figure*}
    \centering
    \includegraphics[width=0.8\linewidth]{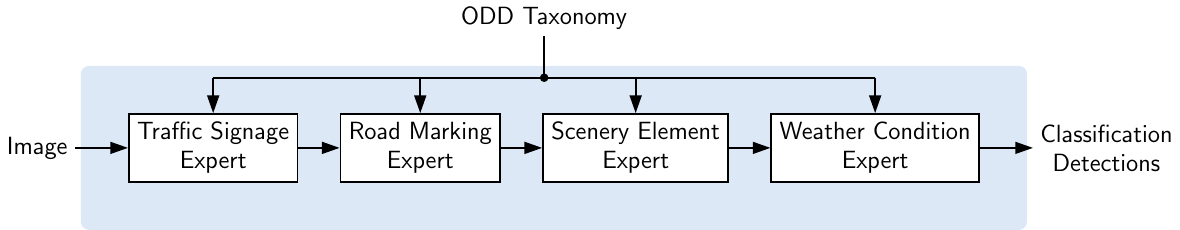}
    \caption{Structure of the chained prompting pipeline as used by \texttt{Persona + Chained CoT} and \texttt{Chained CoT (per-stage heavy)}.}
    \label{fig:chainedpipeline}
\end{figure*}

\textbf{Chain of Thought Prompting (CoT).}
As discussed in~\cref{sec:zeroshotstrategies}, enforcing VLMs to obey structured intermediate reasoning can substantially boost the performance of generative models. 
Specifically, we have an extension of the decomposed persona prompt (\texttt{Persona + CoT}), where we extend the existing prompts via CoT instructions. 
For each domain, the prompt defines criteria that require a brief, specific reasoning script before producing labels. 
For instance, a weather specialist relies on visible evidence, uses defined precipitation intensities (light, moderate, heavy) and ice classifications (graupel, sleet, hail), and follows a decision path: detect precipitation/visibility impairment, identify the main phenomenon from visual cues, assign intensity, and differentiate ice by size/shape. 
If there is uncertainty, it returns up to two plausible labels ranked by confidence.

We also evaluated a chained prompting pipeline that divides detection into sequential stages. 
As visualized in~\cref{fig:chainedpipeline}, the process begins with traffic signage to establish semantic priors, which inform subsequent stages: a road-marking stage infers lane and surface attributes, followed by a scenery stage that integrates signs and markings to define the roadway setting. Next, a weather stage utilizes a CoT approach to identify precipitation and visibility issues, culminating in a trigger-condition stage that considers all prior evidence. This structure leverages dependencies among elements, maintaining strict schema and confidence policies to ensure outputs are machine-ready and auditable. 
In this work, we not only consider the above-described multi-prompt strategy \texttt{Persona + Chained CoT} but also a query-efficient prompt that unifies the entire pipeline in a single prompt (\texttt{Chained CoT (per-stage heavy)}).

\subsection{ODD Taxonomy Element Classification}\label{sec:classification}
%\TODO{ODD taxonomy element Classification}
\noindent Based on the best-performing GPT-4o prompt (see~\cref{sec:resultszeroshot} for details), we utilize~\approach{} to compare different VLMs based on their ODD taxonomy element classification capabilities. 
In the context of an image and a ground-truth ODD element, a VLM accurately \textit{classifies} the image if it predicts the specified ground-truth label for the image. %if it identifies it as an instance of the specified ground-truth ODD element.
In line with our zero-shot optimization study, different models are compared based on their recall against the ground-truth labels. 
We report both the overall recall and the recall for five different element categories (markings, scenery, signs, trigger conditions, and weather).

\textbf{Assessed VLMs.}
Overall, we compare GPT-4o with three types of VLMs: $(i)$ another closed-source, general-purpose multimodal system, $(ii)$ an open-source, general-purpose VLM, and $(iii)$ a smaller, but specialized open-source model. 
For this purpose, we decided to include $(i)$ Gemini 2.5 Pro~\cite{comanici2025gemini} – Google DeepMind's flagship multimodal model designed to handle complex visual and textual reasoning, $(ii)$ Meta AI's Llama 4 Maverick~\cite{meta2025llama4}, which supports long-context visual inputs and excels in tasks involving fine-grained semantic interpretation, and $(iii)$ Molmo~\cite{deitke2025molmo}, a smaller model, specializing in structured reasoning, image captioning, and visual question answering.
For the latter, we use its open-sourced 72B variant.\footnote{\url{https://huggingface.co/allenai/Molmo-72B-0924}}

\subsection{ODD Taxonomy Element Detection}\label{sec:detection}
%\TODO{ODD taxonomy element Detection}
\noindent Beyond merely classifying ODD elements within an image, we further detect the position of observed objects in each image, enabling a downstream fusion of 3D maps from multi-frame camera postures. 
Specifically, given an image, a ground-truth ODD element, and a ground-truth bounding box, we require the model to return the correct label as well as to localize the element using center coordinates.
We additionally report the model's precision, F1-score, and spatial accuracy, which is measured by the average L2 error between the predicted and ground-truth object centers.

\textbf{Detection Specific Dataset.} As outlined in~\cref{sec:prompts}, we designed~\approach{} specifically for evaluating the classification abilities of VLMs (not detection). 
Given the lack of bounding boxes in~\approach{}, we must utilize another existing dataset that, unlike our custom dataset, supports in-image object detection. 
Due to its combination of annotation quality, high-resolution imagery, and environmental diversity, we benchmark our models on the Mapillary Vistas dataset~\cite{8237796}. 
To avoid assessments outside of our ODD taxonomy, we decided to exclude moving objects such as humans, animals, and vehicles.
For this experiment, we adapted the best-performing structured prompting approach from~\cref{sec:resultszeroshot}, with labels organized into four persona groups: construction, environment, road markings, and traffic signs.

\vspace{-.1cm}
\section{Results}

\begin{table}[tbp]
\centering
\caption{Approximate prompt-only token usage per image and corresponding recall by prompting strategy, sorted by GPT-4o recall. \(I\): vision token cost of one image. Token counts are summed across all sub-prompts; strategies with \(5I\) submit the image to each of five domain expert calls separately.}
\label{tab:tokens_per_image_sorted}
\renewcommand{\arraystretch}{1.1}
\begin{tabular}{l r r}
\toprule
\textbf{Prompting Strategy} & \makecell[r]{\textbf{Prompt Tokens} \\ \textbf{per Inference}} & \textbf{Recall (R)} \\
\midrule
Persona + CoT & \(24{,}460 + 5I\) & 0.73 \\
Persona + Chained CoT & \(25{,}980 + 5I\) & 0.68 \\
\makecell[l]{Persona + Label-\\ Aliasing + Taxonomy} & \(12{,}258 + 5I\) & 0.65 \\
Persona Decomposition & \(2{,}117 + 5I\) & 0.59 \\
\makecell[l]{Chained CoT \\ (per-stage heavy)} & \(23{,}600 + I\) & 0.59 \\
Persona + RAG & \(1{,}512 + 5I\) & 0.53 \\
Flat Taxonomy List (baseline) & \(1{,}371 + I\) & 0.40 \\
Reevaluate & \(2{,}793 + 2I\) & 0.36 \\
Road-Dependent & \(372 + 2I\) & 0.22 \\
\bottomrule
\end{tabular}
\end{table}

\vspace{-.1cm}
\subsection{Optimizing Zero-Shot Perception}~\label{sec:resultszeroshot}
\vspace{-.4cm}

\begin{figure}[t!]
  \centering
  \includegraphics[width=\linewidth]{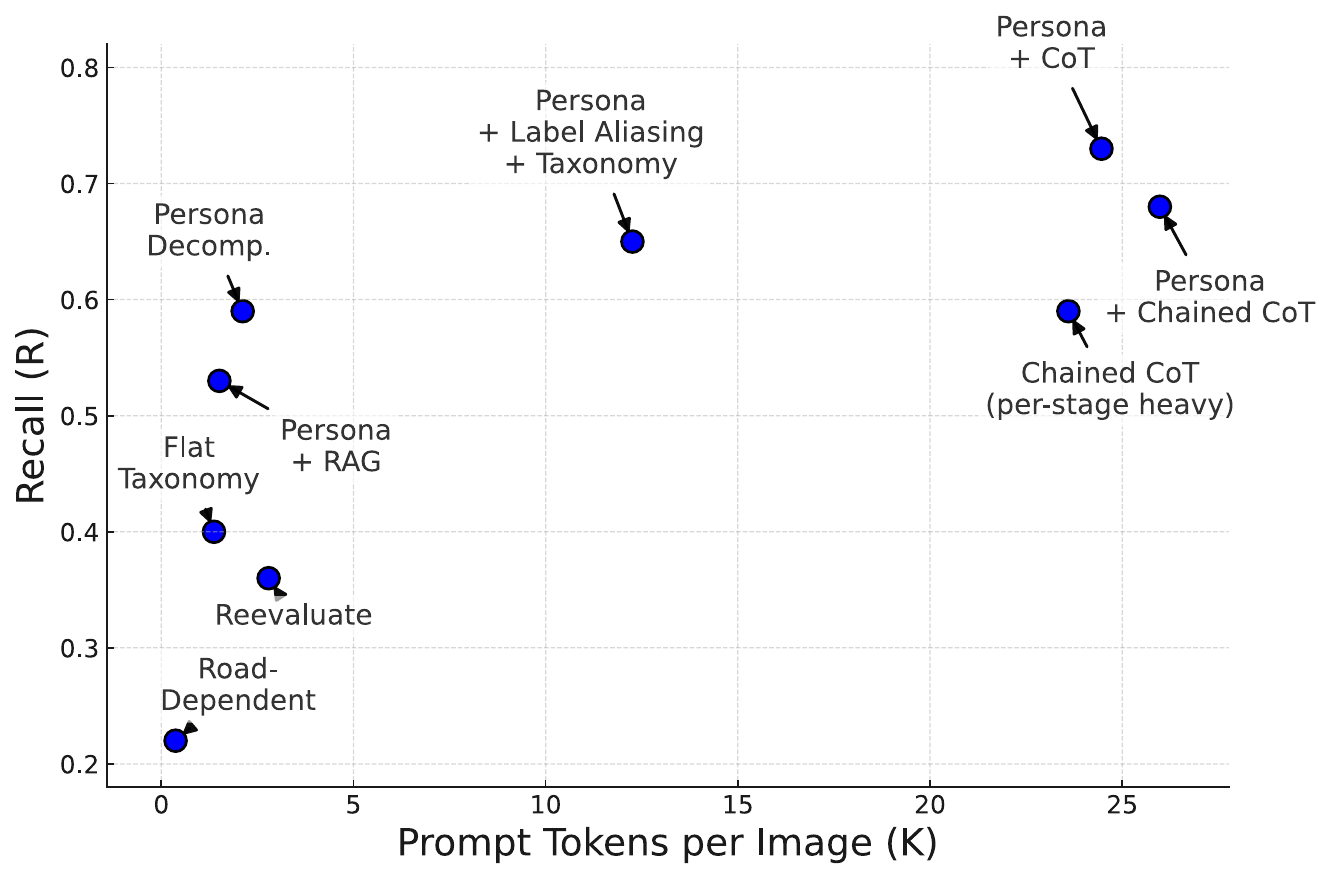}%
  \caption{Performance of different zero-shot optimization strategies in relation to their token budget.}
  \label{fig:tradeoff}
\end{figure}

\noindent Rather than removing components one at a time, each strategy activates a different subset of six design choices: persona 
decomposition, CoT reasoning, taxonomy grouping, label aliasing, RAG retrieval, and chaining, allowing us to observe how component combinations jointly affect recall. The findings of our zero-shot optimization study, along with the corresponding token budgets, are presented in~\cref{tab:tokens_per_image_sorted} and visualized in~\cref{fig:tradeoff}.
These results offer several insights into the effects of zero-shot VLM optimization, which transfer also to more general VLM applications.

\textbf{Token-Efficient Optimization via Persona-Prompting.} 
Despite slightly increasing the token budget per sample, the \texttt{Persona Decomp.} strategy substantially increases the recall ($R = 0.59$) compared to our baseline ($R = 0.4$).

While this strategy involves a higher computational cost due to the requirement for multiple queries, the observed performance boost supports previous empirical findings~\cite{tseng-etal-2024-two,choi2024beyond,luz2025helpful,salewski2023context} on the positive effects of persona-based prompting.

\textbf{Maximizing Performance via CoT.} While the instantiation of personas substantially boosts VLM's zero-shot capabilities, its latent potential is only unlocked when extending the personas with complex chain-of-thought instructions. 
With a recall of $R = 0.73$, our \texttt{Persona + CoT} strategy yields the best performance, confirming that explicit reasoning enhances fine-grained detection. 
However, this configuration and its variants consume up to 25,000 tokens per sample across five personas or stages, posing challenges for throughput and cost in production-scale systems.

\textbf{Avoiding Dependent Multi-Stage Pipelines.}
Among the eight tested zero-shot optimization strategies, three actively harmed VLM perception. 
The strategies \texttt{Reevaluate} and \texttt{Road-dependent} showed significantly poorer performance compared to the flat taxonomy baseline, while incorporating RAG into our persona-based prompt also reduced VLM performance.
These three strategies all share a core design philosophy: a cascaded, upstream-conditioned design that gates later decisions on earlier outputs.
Such a dependent pipeline design is intended to guide the model more effectively by breaking down the task into simpler preliminary steps, such as detecting street types rather than identifying ODD taxonomy concepts right away. 
However, this approach inadvertently creates a mechanism that increases the likelihood of errors in the VLM, as mistakes made in the earlier decisions can propagate through the system.

\begin{table*}[ht]
\centering
\caption{Recall performance of different VLMs on ODD-Tax-232. Best values per column are highlighted in bold.}
\label{tab:odd-recall}
\begin{tabular}{l c c c c c c}
\toprule
\textbf{Model} & \textbf{Overall} & \textbf{Markings} & \textbf{Scenery} & \textbf{Signs} & \textbf{Trigger Conditions} & \textbf{Weather} \\
\midrule
GPT-4o~\cite{hurst2024gpt}               & 0.72 & 0.68 & 0.74 & \textbf{0.74} & 0.69 & 0.72 \\
Maverick~\cite{meta2025llama4}     & 0.67 & 0.62 & 0.68 & 0.67 & 0.65 & \textbf{0.77} \\
Molmo-72B~\cite{deitke2025molmo}         & 0.45 & 0.46 & 0.51 & 0.57 & 0.26 & 0.32 \\
Gemini 2.5 Pro~\cite{comanici2025gemini} & \textbf{0.73} & \textbf{0.72} & \textbf{0.75} & 0.72 & \textbf{0.72} & 0.69 \\
\bottomrule
\end{tabular}
\end{table*}

\subsection{ODD Taxonomy Element Classification}
\noindent Based on the insights gathered from~\cref{sec:resultszeroshot}, we perform all subsequent experiments with the \texttt{Persona-CoT} prompt. 
The resulting zero-shot ODD taxonomy element classification performance of the four VLMs (GPT-4o, Gemini 2.5 Pro, Llama 4 Maverick, and Molmo-72B) is summarized in~\cref{tab:odd-recall}.

\textbf{Large Proprietary vs. Smaller Open Models.}
In terms of performance, we observe a clear separation between the closed-source general-purpose models (GPT-4o and Gemini 2.5 Pro) and their smaller counterparts (Llama 4 Maverick and Molmo-72B).
While the former two models perform on a similar level with a recall of 0.72 and 0.73, respectively, the open-source alternatives demonstrate substantially worse classification capabilities. 
Beyond the overall recall, the closed-source models perform well on signs and coarse road structure and retain a clear edge on fine-grained, safety-critical categories (detailed markings, lane specification, subtle surface damage, scene-level triggers/conditions). 
Llama 4 Maverick is effective with broad geometry and signage but less reliable with rare markings and localized defects. 
In contrast, Molmo-72B struggles with lane counts, damage, and adverse weather conditions. 
These trends align with previous findings, which demonstrate that smaller VLMs underperform on detailed visual reasoning tasks~\cite{chen2025why}.

\begin{figure}[t!]
  \centering
  \includegraphics[width=\linewidth]{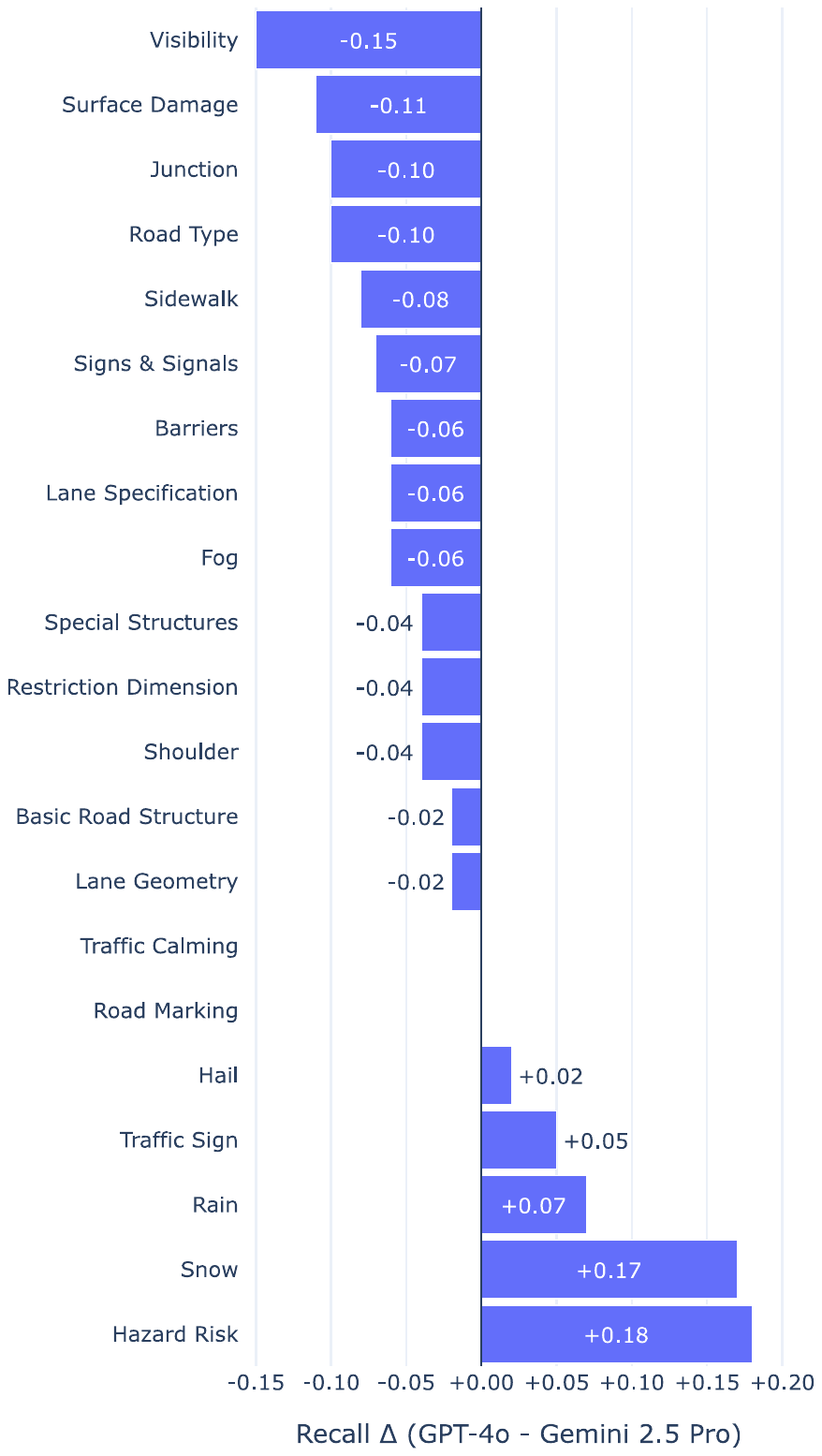}
  \caption{Per-Group Recall Advantage: GPT-4o vs. Gemini 2.5 Pro (Positive = GPT-4o higher recall; Negative = Gemini higher recall).}
  \label{fig:closedsourcecomparison}
\end{figure}

\textbf{GPT-4o vs. Gemini 2.5 Pro.}
In terms of overall recall, Gemini ($0.73$) and GPT-4o ($0.72$) perform almost identically.
However, a per-group recall analysis (cf.~\cref{fig:closedsourcecomparison}) highlights nuanced yet crucial differences:
the performance gains of Gemini are concentrated in markings, scenery, and trigger conditions, whereas GPT-4o retains an edge on signage and weather. 
Hence, the ``better" model depends on the application's risk profile: deployments prioritizing adverse-weather robustness or signage consistency may prefer GPT-4o, whereas settings emphasizing broader scene coverage and lane/marking detail may benefit from Gemini's strengths.
Given the narrow aggregate gap, reporting \emph{both} overall and group-wise metrics is essential to avoid masking domain-specific weaknesses. 
Furthermore, although the prompts were designed for GPT-4o, they achieve a slightly higher recall on Gemini, suggesting that the initial prompt design transfers robustly across models. 
These results are plausibly attributable to differences in pre-training data distribution and instruction-tuning emphasis, which can shift sensitivity toward particular visual taxa (e.g., lane/marking detail versus transient weather phenomena). 
These nuanced differences highlight the importance of taxonomy-aware evaluation, complementing headline metrics.

% Figure

\subsection{ODD Taxonomy Element Detection}

\noindent To complement the evaluation on~\approach{}, we extend the analysis to the Mapillary Vistas dataset. 
As discussed in~\cref{sec:detection}, this dataset enables us to evaluate whether the VLMs can accurately position individual ODD taxonomy concepts within a single image.

The evaluation results (cf.~\cref{tab:mapillary-eval}) show that GPT-4o achieves the best overall performance, with the highest precision (0.82) and F1-score (0.87). 
This indicates that GPT-4o minimizes false positives while maintaining a strong balance with recall, which is crucial for safety-critical applications. 
Gemini 2.5 Pro achieves the highest recall ($0.977$), detecting more ground-truth objects, but at the expense of precision ($0.71$), resulting in a higher rate of false positives and a lower F1-score than GPT-4o. 
Maverick demonstrates the best spatial accuracy, with the lowest Average L2 Error of $0.11$, indicating that when it detects accurately, its position estimates are highly precise.
However, its overall F1-score ($0.85$) falls behind that of GPT-4o.
Finally, Molmo-72B lags significantly in precision ($0.46$) and F1-score ($0.621$), making it less reliable for this task.

Overall, while Gemini 2.5 Pro excels in recall and Maverick in spatial accuracy, GPT-4o strikes the most favorable balance between detection reliability and precision. 
For the static taxonomy element detection task, this balance is paramount: in ODD monitoring, a false positive flags a road segment as non-compliant and triggers an unnecessary safe fallback or disengagement, which in dense traffic is itself a safety hazard, making precision a critical deployment consideration alongside recall.
Therefore, we identify GPT-4o as the most robust and dependable VLM for Mapillary Vistas object detection.

\begin{table}[ht]
\centering
\caption{Evaluation of VLMs on the Mapillary Vistas dataset. Best values per column are highlighted in bold. Average L2 Error is the mean Euclidean distance between the predicted point and the ground-truth box center for matched detections (lower is better).}
\label{tab:mapillary-eval}
\begin{tabular}{l c c c c}
\toprule
\textbf{Model} & \textbf{Precision} & \textbf{Recall} & \textbf{F1-Score} & \makecell{\textbf{Average} \\ \textbf{L2 Error}} \\
\midrule
GPT-4o~\cite{hurst2024gpt}        & \textbf{0.82} & 0.922 & \textbf{0.87}  & 0.18 \\
Maverick~\cite{meta2025llama4} & 0.80          & 0.914 & 0.85           & \textbf{0.11} \\
Molmo-72B~\cite{deitke2025molmo} & 0.46          & 0.937 & 0.621          & 0.20 \\
Gemini 2.5 Pro~\cite{comanici2025gemini} & 0.71          & \textbf{0.977} & 0.821          & 0.14 \\
\bottomrule
\end{tabular}
\end{table}

\noindent \textbf{Discussion: ODD-TAX-232 vs. Mapillary Vistas.}
Across both datasets, all VLMs score higher on Mapillary Vistas than on~\approach{}.
The main reason is simple: Mapillary Vistas has a much smaller, coarser set of element types that are common in training data and easy to interpret visually (e.g., roads, buildings, standard signs). 
Its dense, mature annotations further reduce noise and stabilize evaluation. 
By contrast,~\approach{} targets many more, fine-grained ODD elements, including rare, safety-relevant cases and subtle cues, which are less frequent in pretraining corpora and therefore harder for current VLMs.
% This gap reflects the finer-grained, composite labels in~\approach{}, which demands multi-cue, relational reasoning, whereas many Mapillary classes are visually salient and atomic. 
% Mapillary's dense, mature annotations and head-class prevalence align well with VLM priors. 
% At the same time,~\approach{} emphasizes rare, safety-relevant edge cases and subtle, context-dependent cues --- both of which depress recall under single-frame prompting.
% A further contributor may be distributional overlap: some VLMs likely encountered Mapillary-like imagery during pre-training, which aids transfer more than our custom ODD taxonomy.

Higher Mapillary scores do not contradict the results on~\approach{}. 
Instead, they indicate that current VLMs are stronger on frequent, visually salient categories than on fine-grained, safety-critical concepts requiring compositional reasoning.
Thus, harnessing the full potential of VLMs for ODD perception requires fine-tuning for uncommon, taxonomy-specific element types.
Consequently, aggregate metrics on Mapillary can overestimate readiness for ODD perception unless accompanied by taxonomy-aware analyses.

\section{Conclusion}

\noindent In this work, we explore the use of VLMs for extracting complex Operational Design Domain (ODD) taxonomy concepts from camera imagery in autonomous driving. 
We evaluate these models using a custom ODD dataset (\approach{}) and experiments on Mapillary Vistas, with a focus on their current zero-shot perception capabilities.

Our findings lead to three main conclusions.
% First, while VLMs show potential for ODD enrichment, they currently fall short of meeting the demands of autonomous driving, particularly for subtle and edge cases. 
First, VLMs show promising potential for ODD conditions recognition in camera images. However, from a safety-criticality perspective, a recall of $R = 0.73$ is insufficient for standalone production deployment as an on-board ODD sensor, where undetected ODD violations could allow system activation under unsafe conditions. Nevertheless, this level of performance positions VLMs as a viable tool for offline applications, such as pre-route ODD auditing, regulatory compliance reporting, and simulation seeding, where the cost of missed detections is lower, and human oversight can compensate for residual errors.

Second, the design of prompts significantly impacts VLM performance, with tailored CoT prompting with elaborated low-level reasoning steps yielding the best balance of coverage and output quality.
Third, comparisons among models (GPT-4o, Gemini 2.5 Pro, Maverick, Molmo) revealed small differences in overall recall but highlighted the need for taxonomy-aware evaluations.

Future work may transfer our zero-shot optimization templates to other domains, such as maritime systems and agricultural robots.
Additionally, exploring novel architectures (such as GPT-5) and enhancing ODD conditions perception through data fusion and the development of a taxonomy-aware, open-source VLM are promising avenues for advancement. 
Overall, our findings support the notion that VLMs can significantly contribute to ODD conditions enrichment, particularly when combined with focused prompting and normalization techniques.

\section{DISCLAIMER}
\noindent The results, opinions, and conclusions expressed in this publication are not necessarily those of Volkswagen Aktiengesellschaft.

\bibliographystyle{IEEEtran}
\bibliography{references.bib}

@INPROCEEDINGS{10920066,
  author={Sural, Shounak and Naren and Rajkumar, Ragunathan Raj},
  booktitle={2024 IEEE 27th International Conference on Intelligent Transportation Systems (ITSC)}, 
  title={ContextVLM: Zero-Shot and Few-Shot Context Understanding for Autonomous Driving Using Vision Language Models}, 
  year={2024},
  volume={},
  number={},
  pages={468-475},
  doi={10.1109/ITSC58415.2024.10920066}}

@article{li2025vladbench,
  title   = {Fine-Grained Evaluation of Large Vision-Language Models in Autonomous Driving},
  author  = {Li, Yue and Tian, Meng and Lin, Zhenyu and Zhu, Jiangtong and Zhu, Dechang and Liu, Haiqiang and Wang, Zining and Zhang, Yueyi and Xiong, Zhiwei and Zhao, Xinhai},
  journal = {arXiv preprint arXiv:2503.21505},
  year    = {2025}
}

@article{gao2025foundation,
  author    = {Gao, Yuan and others},
  title     = {Foundation Models in Autonomous Driving: A Survey on Scenario Generation and Scenario Analysis},
  journal   = {arXiv preprint arXiv:2506.11526},
  year      = {2025}
}

@article{hou2024vlmADASurvey,
  author  = {Zhou, Xingcheng and Liu, Mingyu and Yurtsever, Ekim and {\v{Z}}agar, Bare Luka and Zimmer, Walter and Cao, Hu and Knoll, Alois C.},
  title   = {Vision{-}Language Models in Autonomous Driving: A Survey and Outlook},
  journal = {IEEE Transactions on Intelligent Vehicles},
  year    = {2024},
  pages   = {1--20},
  doi     = {10.1109/TIV.2024.3402136},
  url     = {https://doi.org/10.1109/TIV.2024.3402136},
  note    = {Early Access}
}

@misc{smith2024finegrainedvl,
      title={LEGO Co-builder: Exploring Fine-Grained Vision-Language Modeling for Multimodal LEGO Assembly Assistants}, 
      author={Haochen Huang and Jiahuan Pei and Mohammad Aliannejadi and Xin Sun and Moonisa Ahsan and Chuang Yu and Zhaochun Ren and Pablo Cesar and Junxiao Wang},
      year={2025},
      eprint={2507.05515},
      archivePrefix={arXiv},
      primaryClass={cs.AI},
      url={https://arxiv.org/abs/2507.05515}, 
}

@inproceedings{marathe2023wedge,
  author    = {Marathe, Aboli and Ramanan, Deva and Walambe, Rahee and Kotecha, Ketan},
  title     = {{WEDGE}: A multi-weather autonomous driving dataset built from generative vision-language models},
  booktitle = {Proceedings of the IEEE/CVF Conference on Computer Vision and Pattern Recognition (CVPR) Workshops},
  year      = {2023}
}

@article{lu2024omnitester,
  author    = {Lu, Qiujing and Wang, Xuanhan and Jiang, Yiwei and Zhao, Guangming and Ma, Mingyue and Feng, Shuo},
  title     = {Multimodal Large Language Model Driven Scenario Testing for Autonomous Vehicles},
  journal   = {arXiv preprint arXiv:2409.06450},
  year      = {2024}
}

@misc{zhang2025safeautoknowledgeenhancedsafeautonomous,
      title={SafeAuto: Knowledge-Enhanced Safe Autonomous Driving with Multimodal Foundation Models}, 
      author={Jiawei Zhang and Xuan Yang and Taiqi Wang and Yu Yao and Aleksandr Petiushko and Bo Li},
      year={2025},
      eprint={2503.00211},
      archivePrefix={arXiv},
      primaryClass={cs.RO},
      url={https://arxiv.org/abs/2503.00211}, 
}

@INPROCEEDINGS{8237796,
  author={Neuhold, Gerhard and Ollmann, Tobias and Bulò, Samuel Rota and Kontschieder, Peter},
  booktitle={2017 IEEE International Conference on Computer Vision (ICCV)}, 
  title={The Mapillary Vistas Dataset for Semantic Understanding of Street Scenes}, 
  year={2017},
  volume={},
  number={},
  pages={5000-5009},
  doi={10.1109/ICCV.2017.534}}

@article{hancock2019future,
  title={On the future of transportation in an era of automated and autonomous vehicles},
  author={Hancock, Peter A and Nourbakhsh, Illah and Stewart, Jack},
  journal={Proceedings of the National Academy of Sciences},
  volume={116},
  number={16},
  pages={7684--7691},
  year={2019},
  publisher={National Academy of Sciences}
}

@standard{ISO34503_2023,
  author       = {{International Organization for Standardization}},
  title        = {{ISO 34503:2023 - Road vehicles - Test scenarios for automated driving systems - Taxonomy for Operational Design Domain (ODD)}},
  year         = {2023},
  publisher    = {International Organization for Standardization},
  institution  = {{International Organization for Standardization}},
  address      = {Geneva, Switzerland},
  note        = {Available from \url{https://www.iso.org/standard/78952.html}}
}

@standard{ISO21448_2022,
  author      = {{International Organization for Standardization}},
  title       = {{ISO 21448:2022 - Road vehicles — Safety of the intended functionality}},
  year        = {2022},
  publisher   = {International Organization for Standardization},
  institution  = {{International Organization for Standardization}},
  address     = {Geneva, Switzerland},
  note        = {Available from \url{https://www.iso.org/standard/70939.html}}
}

@misc{BSI2020PAS1883,
  author       = {{The British Standards Institution}},
  title        = {PAS 1883 - Operational Design Domain (ODD) Taxonomy for ADS Specification},
  year         = {2020},
  publisher    = {BSI Standards Limited 2020},
  isbn         = {978 0 539 06735 4},
  note         = {Published by BSI Standards Limited. First published August 2020. No copying without BSI permission except as permitted by copyright law.},
  howpublished = {Available from BSI Group},
  address      = {London, United Kingdom},
  issn         = {ICS 03.220.20; 35.240.60}
}

@inproceedings{gyllenhammar2020towards,
  title={Towards an operational design domain that supports the safety argumentation of an automated driving system},
  author={Gyllenhammar, Magnus and Johansson, Rolf and Warg, Fredrik and Chen, DeJiu and Heyn, Hans-Martin and Sanfridson, Martin and S{\"o}derberg, Jan and Thors{\'e}n, Anders and Ursing, Stig},
  booktitle={10th European congress on embedded real time systems (ERTS 2020)},
  year={2020}
}

@misc{ISO34505_2025,
  author       = {{International Organization for Standardization}},
  title        = {{ISO 34505:2025 -- Road vehicles -- Test scenarios for automated driving systems -- Scenario evaluation and test case generation}},
  year         = {2025},
  howpublished = {International Standard},
  publisher    = {ISO},
  address      = {Geneva, Switzerland},
  url          = {https://www.iso.org/standard/78954.html},
  note         = {Accessed 2025-08-14}
}

@article{broggi2015proud,
  title={Proud—public road urban driverless-car test},
  author={Broggi, Alberto and Cerri, Pietro and Debattisti, Stefano and Laghi, Maria Chiara and Medici, Paolo and Molinari, Daniele and Panciroli, Matteo and Prioletti, Antonio},
  journal={IEEE Transactions on Intelligent Transportation Systems},
  volume={16},
  number={6},
  pages={3508--3519},
  year={2015},
  publisher={IEEE}
}

@inproceedings{NEURIPS2020_1457c0d6,
 author = {Brown, Tom and Mann, Benjamin and Ryder, Nick and Subbiah, Melanie and Kaplan, Jared D and Dhariwal, Prafulla and Neelakantan, Arvind and Shyam, Pranav and Sastry, Girish and Askell, Amanda and Agarwal, Sandhini and Herbert-Voss, Ariel and Krueger, Gretchen and Henighan, Tom and Child, Rewon and Ramesh, Aditya and Ziegler, Daniel and Wu, Jeffrey and Winter, Clemens and Hesse, Chris and Chen, Mark and Sigler, Eric and Litwin, Mateusz and Gray, Scott and Chess, Benjamin and Clark, Jack and Berner, Christopher and McCandlish, Sam and Radford, Alec and Sutskever, Ilya and Amodei, Dario},
 booktitle = {Advances in Neural Information Processing Systems},
 editor = {H. Larochelle and M. Ranzato and R. Hadsell and M.F. Balcan and H. Lin},
 pages = {1877--1901},
 publisher = {Curran Associates, Inc.},
 title = {Language Models are Few-Shot Learners},
 url = {https://proceedings.neurips.cc/paper_files/paper/2020/file/1457c0d6bfcb4967418bfb8ac142f64a-Paper.pdf},
 volume = {33},
 year = {2020}
}

@inproceedings{10.5555/3600270.3602070,
author = {Wei, Jason and Wang, Xuezhi and Schuurmans, Dale and Bosma, Maarten and Ichter, Brian and Xia, Fei and Chi, Ed H. and Le, Quoc V. and Zhou, Denny},
title = {Chain-of-thought prompting elicits reasoning in large language models},
year = {2022},
isbn = {9781713871088},
publisher = {Curran Associates Inc.},
address = {Red Hook, NY, USA},
booktitle = {Proceedings of the 36th International Conference on Neural Information Processing Systems},
articleno = {1800},
numpages = {14},
location = {New Orleans, LA, USA},
series = {NIPS '22}
}

@article{zhang2022automatic,
  title={Automatic chain of thought prompting in large language models},
  author={Zhang, Zhuosheng and Zhang, Aston and Li, Mu and Smola, Alex},
  journal={arXiv preprint arXiv:2210.03493},
  year={2022}
}

@misc{wang2023selfconsistencyimproveschainthought,
      title={Self-Consistency Improves Chain of Thought Reasoning in Language Models}, 
      author={Xuezhi Wang and Jason Wei and Dale Schuurmans and Quoc Le and Ed Chi and Sharan Narang and Aakanksha Chowdhery and Denny Zhou},
      year={2023},
      eprint={2203.11171},
      archivePrefix={arXiv},
      primaryClass={cs.CL},
      url={https://arxiv.org/abs/2203.11171}, 
}

@article{lewis2020retrieval,
  title={Retrieval-augmented generation for knowledge-intensive nlp tasks},
  author={Lewis, Patrick and Perez, Ethan and Piktus, Aleksandra and Petroni, Fabio and Karpukhin, Vladimir and Goyal, Naman and K{\"u}ttler, Heinrich and Lewis, Mike and Yih, Wen-tau and Rockt{\"a}schel, Tim and others},
  journal={Advances in neural information processing systems},
  volume={33},
  pages={9459--9474},
  year={2020}
}

@inproceedings{dhuliawala2024chain,
  title={Chain-of-verification reduces hallucination in large language models},
  author={Dhuliawala, Shehzaad and Komeili, Mojtaba and Xu, Jing and Raileanu, Roberta and Li, Xian and Celikyilmaz, Asli and Weston, Jason},
  booktitle={Findings of the association for computational linguistics: ACL 2024},
  pages={3563--3578},
  year={2024}
}

@inproceedings{yu2024chain,
  title={Chain-of-note: Enhancing robustness in retrieval-augmented language models},
  author={Yu, Wenhao and Zhang, Hongming and Pan, Xiaoman and Cao, Peixin and Ma, Kaixin and Li, Jian and Wang, Hongwei and Yu, Dong},
  booktitle={Proceedings of the 2024 conference on empirical methods in natural language processing},
  pages={14672--14685},
  year={2024}
}

@article{10.1145/3560815,
author = {Liu, Pengfei and Yuan, Weizhe and Fu, Jinlan and Jiang, Zhengbao and Hayashi, Hiroaki and Neubig, Graham},
title = {Pre-train, Prompt, and Predict: A Systematic Survey of Prompting Methods in Natural Language Processing},
year = {2023},
issue_date = {September 2023},
publisher = {Association for Computing Machinery},
address = {New York, NY, USA},
volume = {55},
number = {9},
issn = {0360-0300},
url = {https://doi.org/10.1145/3560815},
doi = {10.1145/3560815},
journal = {ACM Comput. Surv.},
month = jan,
articleno = {195},
numpages = {35}
}

@article{hurst2024gpt,
  title={Gpt-4o system card},
  author={Hurst, Aaron and Lerer, Adam and Goucher, Adam P and Perelman, Adam and Ramesh, Aditya and Clark, Aidan and Ostrow, AJ and Welihinda, Akila and Hayes, Alan and Radford, Alec and others},
  journal={arXiv preprint arXiv:2410.21276},
  year={2024}
}

@inproceedings{deitke2025molmo,
  title={Molmo and pixmo: Open weights and open data for state-of-the-art vision-language models},
  author={Deitke, Matt and Clark, Christopher and Lee, Sangho and Tripathi, Rohun and Yang, Yue and Park, Jae Sung and Salehi, Mohammadreza and Muennighoff, Niklas and Lo, Kyle and Soldaini, Luca and others},
  booktitle={Proceedings of the Computer Vision and Pattern Recognition Conference},
  pages={91--104},
  year={2025}
}

@article{comanici2025gemini,
  title={Gemini 2.5: Pushing the frontier with advanced reasoning, multimodality, long context, and next generation agentic capabilities},
  author={Comanici, Gheorghe and Bieber, Eric and Schaekermann, Mike and Pasupat, Ice and Sachdeva, Noveen and Dhillon, Inderjit and Blistein, Marcel and Ram, Ori and Zhang, Dan and Rosen, Evan and others},
  journal={arXiv preprint arXiv:2507.06261},
  year={2025}
}

@misc{mckinsey2023autonomous,
  author = {{McKinsey \& Company}},
  title = {Autonomous driving's future: Convenient and connected},
  year = {2023},
  month = {January},
  url = {https://www.mckinsey.com/industries/automotive-and-assembly/our-insights/autonomous-drivings-future-convenient-and-connected},
  note = {Accessed: 2025-09-01}
}

@article{nakashima2025maritime,
journal={Reliability Engineering and System Safety},
author={Nakashima, Takuya and Kureta, Rui and Khastgir, Siddartha},
title={Addressing systemic risks in autonomous maritime navigation: A structured STPA and ODD-based methodology},
year={2025},
volume={261},
number={C},
doi={10.1016/j.ress.2025.111041},
url={https://ideas.repec.org/a/eee/reensy/v261y2025ics095183202500242x.html},
}

@inproceedings{torens2024operational,
  title={From Operational Design Domain to Runtime Monitoring of AI-Based Aviation Systems},
  author={Torens, Christoph and Gupta, Siddhartha and Roy, Nirmal and Sprockhoff, Jasper and Durak, Umut},
  booktitle={2024 AIAA DATC/IEEE 43rd Digital Avionics Systems Conference (DASC)},
  pages={1--9},
  year={2024},
  organization={IEEE}
}

@article{Matin2024publicperception, 
author = {Ali Matin and Hussein Dia},
title = {Public perception of connected and automated vehicles: Benefits, concerns, and barriers from an Australian perspective},
year = {2024},
journal = {Journal of Intelligent and Connected Vehicles},
volume = {7},
number = {2},
pages = {108-128},
url = {https://www.sciopen.com/article/10.26599/JICV.2023.9210028},
doi = {10.26599/JICV.2023.9210028}
}

@misc{kba2023,
	title = {{Nationale} {Betriebserlaubnis} für {Kraftfahrzeuge} mit autonomer {Fahrfunktion}},
	url = {https://www.kba.de/DE/Themen/Typgenehmigung/Autonomes_automatisiertes_Fahren/nationale_Betriebserlaubnis/nationale_betriebserlaubnis_node.html},
	urldate = {2025-09-02},
    author = {Kraftfahrt-{Bundesamt}},
    year = 2023
}

@book{sae2014odd,
  title={Taxonomy and Definitions for Terms Related to Driving Automation Systems for On-Road Motor Vehicles (J3016\_202104)},
  author={{On-Road Automated Driving (ORAD) Committee}},
  year={2014},
  publisher={SAE International}
}

@inproceedings{tseng-etal-2024-two,
    title = "Two Tales of Persona in {LLM}s: A Survey of Role-Playing and Personalization",
    author = "Tseng, Yu-Min  and
      Huang, Yu-Chao  and
      Hsiao, Teng-Yun  and
      Chen, Wei-Lin  and
      Huang, Chao-Wei  and
      Meng, Yu  and
      Chen, Yun-Nung",
    editor = "Al-Onaizan, Yaser  and
      Bansal, Mohit  and
      Chen, Yun-Nung",
    booktitle = "Findings of the Association for Computational Linguistics: EMNLP 2024",
    month = nov,
    year = "2024",
    address = "Miami, Florida, USA",
    publisher = "Association for Computational Linguistics",
    url = "https://aclanthology.org/2024.findings-emnlp.969/",
    doi = "10.18653/v1/2024.findings-emnlp.969",
    pages = "16612--16631"
}

@inproceedings{choi2024beyond,
      title={PICLe: Eliciting Diverse Behaviors from Large Language Models with Persona In-Context Learning}, 
      author={Hyeong Kyu Choi and Yixuan Li},
      booktitle={International Conference on Machine Learning},
      year={2024}
}

@article{luz2025helpful,
  title={Helpful assistant or fruitful facilitator? Investigating how personas affect language model behavior},
  author={Luz de Araujo, Pedro Henrique and Roth, Benjamin},
  journal={PloS one},
  volume={20},
  number={6},
  pages={e0325664},
  year={2025},
  publisher={Public Library of Science San Francisco, CA USA}
}

@article{salewski2023context,
  title={In-context impersonation reveals large language models' strengths and biases},
  author={Salewski, Leonard and Alaniz, Stephan and Rio-Torto, Isabel and Schulz, Eric and Akata, Zeynep},
  journal={Advances in neural information processing systems},
  volume={36},
  pages={72044--72057},
  year={2023}
}

@misc{meta2025llama4,
  title        = {Introducing LLaMA 4: Advancing Multimodal Intelligence},
  author       = {{Meta AI}},
  year         = {2025},
  howpublished = {Meta AI Blog},
  url          = {https://ai.meta.com/blog/llama-4-multimodal-intelligence/},
  note         = {Accessed: 2025-09-08}
}

@inproceedings{chen2025why,
  title={Why Is Spatial Reasoning Hard for VLMs? An Attention Mechanism Perspective on Focus Areas},
  author={Chen, Shiqi and Zhu, Tongyao and Zhou, Ruochen and Zhang, Jinghan and Gao, Siyang and Niebles, Juan Carlos and Geva, Mor and He, Junxian and Wu, Jiajun and Li, Manling},
  booktitle={International Conference on Machine Learning},
  pages={9910--9932},
  year={2025},
  organization={PMLR}
}

\end{document}